%% file: iclr2027_conference.tex
\documentclass{article} 
\usepackage{iclr2027_conference,times}

\input{math_commands.tex}

\usepackage{hyperref}
\usepackage{cleveref}
\usepackage{url}
\usepackage{booktabs}
\usepackage{multirow}
\usepackage{graphicx}
\usepackage{wrapfig}
\usepackage[most]{tcolorbox}

\crefname{section}{Sec.}{Secs.}
\Crefname{section}{Sec.}{Secs.}
\crefname{subsection}{Sec.}{Secs.}
\Crefname{subsection}{Sec.}{Secs.}
\crefname{figure}{Fig.}{Figs.}
\Crefname{figure}{Fig.}{Figs.}

\usepackage[table]{xcolor}
\usepackage{xspace}
\newcommand{\method}{\textsc{Wedge}\xspace}

\title{From Failures to Learnable Worlds: Policy-Conditioned Environment Generation for Navigation Agents
}

\author{
Yue Zhang \quad
Han Lin \quad
Zun Wang \quad
Yidong Huang \quad
Zaid Khan \quad
Mohit Bansal
\\[0.5em]
\normalfont UNC Chapel Hill
}

\usepackage[table]{xcolor}
\definecolor{oursbg}{RGB}{243,237,250}
\newcommand{\ourscell}{\cellcolor{oursbg}}

\iclrfinalcopy 
\begin{document}

\maketitle

\input{sections/abstract}
\input{sections/intro}

\input{sections/related_work}

\input{sections/method}

\input{sections/experiment}

\input{sections/conclusion}

\bibliography{iclr2027_conference}
\bibliographystyle{iclr2027_conference}

\appendix
\input{sections/appendix}

\end{document}

%% file: math_commands.tex
\usepackage{amsmath,amsfonts,bm}

\def\eqref#1{equation~\ref{#1}}

\def\1{\bm{1}}

\DeclareMathAlphabet{\mathsfit}{\encodingdefault}{\sfdefault}{m}{sl}
\SetMathAlphabet{\mathsfit}{bold}{\encodingdefault}{\sfdefault}{bx}{n}



%% file: sections/abstract.tex
\begin{abstract}

Simulated environments enable scalable training data generation for embodied agents, but more data does not necessarily provide what a particular agent needs to learn. Existing environment-generation pipelines largely optimize generic properties such as realism, diversity, or task coverage, while different pretrained policies can fail for different environmental reasons, with some struggling when targets are hard to see and others when navigation requires long blind paths or multiple room transitions.
We therefore introduce \method{}, a framework for weakness-guided navigation learning in which the failure modes of a pretrained agent guide the generation of training environments at each adaptation round.
Rather than assuming a predefined set of tasks or skills to improve, \method{} derives its generation targets from the policy's own success and failure patterns within the same task.
Given a pretrained policy, \method{} uses an LLM to propose executable weakness features from contrasting successful and failed rollouts, automatically validates them on the diagnostic set, and selects the strongest validated weakness based on its success-rate gap.
The validated weaknesses are then mapped to the environmental factors that control them, such as target placement, start configurations, and scene layouts. \method{} varies these factors to construct environments in the difficulty regimes where the policy struggles, producing targeted training experience for adaptation.
Across three navigation agents with different adaptation mechanisms, we find
that policies exhibit distinct weakness profiles and benefit from different
training environments.
Weakness-conditioned adaptation consistently outperforms matched random and
hard-example alternatives on ProcTHOR, while the resulting gains for Uni-NaVid
and Gemini also extend to HM3D after ProcTHOR adaptation.
Iterative re-diagnosis further shows that dominant weaknesses can persist or
shift as the policy improves, providing updated guidance for subsequent
environment construction.
Together, these results show that policy failures are not only diagnostic
signals, but actionable guidance for constructing training environments that
target a policy's current weaknesses
\footnote{%
\begin{tabular}[t]{@{}l@{\ }l@{}}
Website: & \url{https://zhangyuejoslin.github.io/WEDGE} \\
Code:    & \url{https://github.com/zhangyuejoslin/WEDGE}
\end{tabular}}.

\end{abstract}

%% file: sections/intro.tex
\section{Introduction}
Simulated environments have become an important source of scalable training data for embodied agents~\citep{wang2025embodiedgen, ye20263d, seed2025seed3d, yang2024holodeck, deitke2022, chang2017matterport3d, ramakrishnan2021hm3d}, offering controllable worlds, inexpensive interaction, and access to situations that are difficult to collect in the real world. Accordingly, recent environment-generation pipelines have progressed from improving realism and diversity~\citep{yang2024holodeck,pan2026metaspatial,sun2025layoutvlm, hu2024scenecraft} to task-conditioned generation, where environments are constructed to support specific embodied tasks~\citep{xia2026sage, yang2026sceneweaver, wang2026embodiedgen,zala2024envgen}. However, these approaches primarily condition generation on predefined properties of the environment or user-specified embodied tasks (e.g., ``Find the mug.''), with limited consideration of what the current policy has already mastered and where it still struggles.
As a result, the same training environment may be useful for one policy but redundant or overly difficult for another. This raises a central question: \emph{how can we generate environments that target what a particular agent has not yet mastered?}

\begin{figure}
    \centering
    \includegraphics[width=\linewidth]{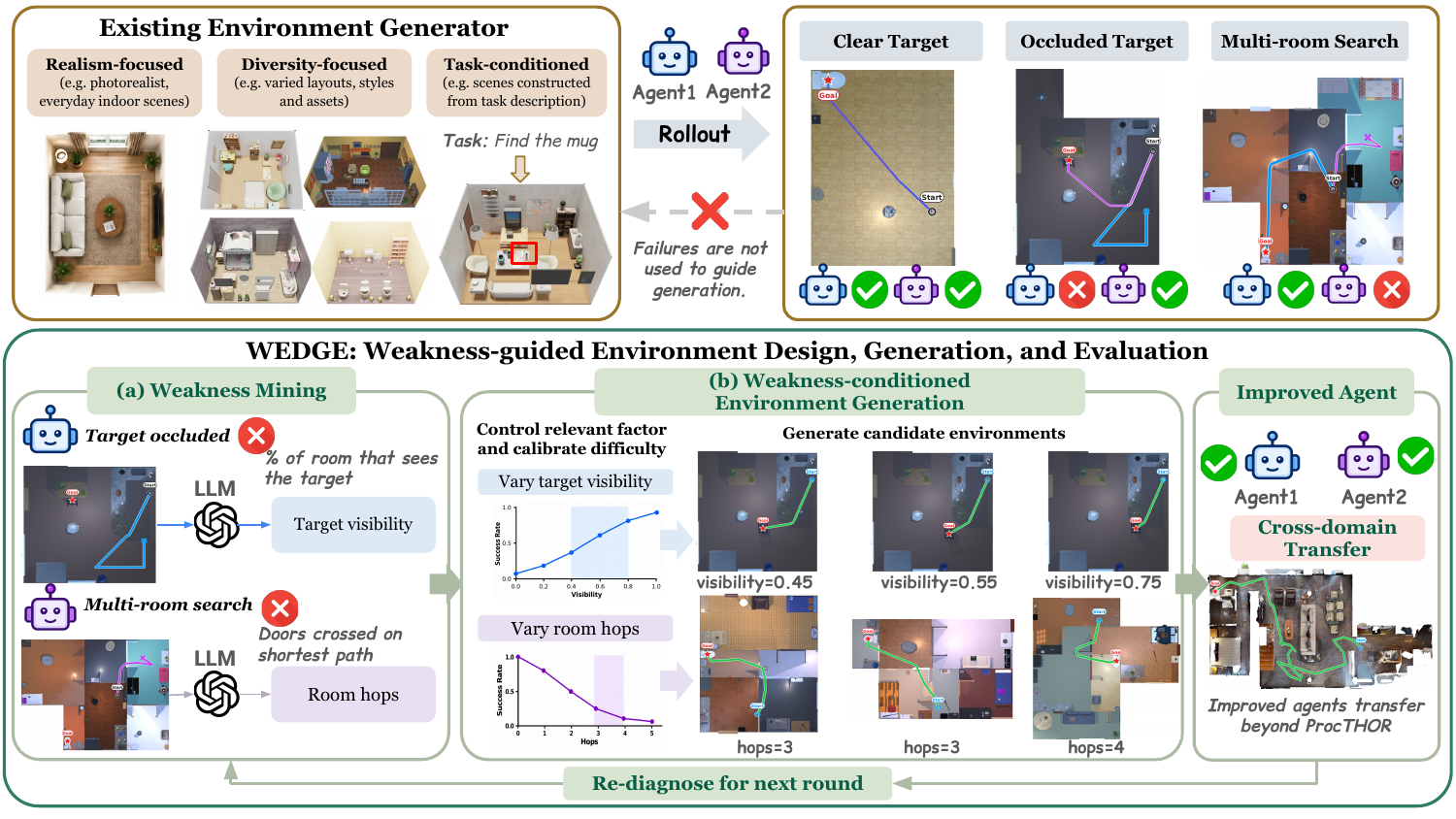}
    \caption{\textbf{Overview of \method.}
Unlike existing environment generators, which are largely policy-agnostic, \method uses agent failures to guide generation.
(a) We mine policy-specific weaknesses from successful and failed rollouts.
(b) The mined weaknesses determine which environmental factors to control and which learnable regimes to target.
Training on the resulting environments improves the agent and can yield cross-domain transfer beyond the generation simulator.
}
    \label{fig:teaser_figure}
\end{figure}

This distinction is particularly important when continuing to train a pretrained embodied agent, whose capabilities may already vary substantially across environmental conditions.
The same training environment may therefore be redundant for an agent that has already mastered the relevant behavior, informative for another that still struggles with it, or too difficult to provide a useful learning signal. As illustrated in~\Cref{fig:teaser_figure} (top), two agents solving the same navigation task can fail for very different reasons: one may struggle when navigation requires traversing multiple rooms, while another may fail primarily when the target is poorly visible. 
Generating environments that are simply more realistic, diverse, or difficult does not necessarily address either weakness. Task-conditioned generation provides a stronger form of control by constructing environments around the requirements of a given task, but it does not distinguish why different policies fail within that task. As a result, environments generated for the same task may still target conditions that one policy has already mastered while overlooking those that another policy struggles with. We therefore argue that training environments for continued learning should be generated relative to the current policy.

We introduce \textbf{\method{}}
(\textbf{W}eakness-guided \textbf{E}nvironment \textbf{D}esign,
\textbf{G}eneration, and \textbf{E}valuation), a framework for procedural
environment generation that closes the loop between policy diagnosis,
environment construction, and policy adaptation for object navigation tasks.
Given a pretrained navigation policy, we first contrast its successful and failed episodes to identify environmental factors that systematically characterize where the policy struggles. An LLM instantiates these hypotheses as programmatic weakness features, such as \texttt{target\_visibility}, \texttt{blind\_path\_ratio}, and \texttt{room\_hops}, which can be computed over individual environments and directly connected to controllable simulator factors (\Cref{fig:teaser_figure}(a)). We evaluate each proposed feature on the full diagnostic set and retain only those that provide sufficiently broad, non-redundant, predictive, and complementary signals of policy failure. These validated weaknesses then serve as actionable specifications for generation, determining which environmental factors should be varied and which failure-associated regimes should be targeted (\Cref{fig:teaser_figure}(b)). 
A key aspect of \method{} is that its generation targets are derived from the current policy's own failure patterns rather than specified in advance. By identifying the environmental conditions associated with failures within the same task, \method{} determines which aspects of the environment should be modified and which regimes should be targeted during generation.
We then roll out the current policy in each candidate environment and select those that fall within the diagnosed weak regime while remaining challenging but solvable. The selected environments are then used to adapt the policy with its original learning mechanism.
We study this framework in 3D indoor navigation, where agents must continuously perceive, navigate, and act in complex environments over long horizons. We further show that weakness-targeted adaptation in ProcTHOR~\citep{deitke2022} transfers to HM3D~\citep{ramakrishnan2021habitat}, demonstrating generalization beyond the simulator used for generation.

We evaluate \method{} across three navigation agents with different model architectures and adaptation mechanisms: a ProcTHOR~\citep{deitke2022}-trained DD-PPO policy~\citep{zeng2024poliformer}, an HM3D~\citep{ramakrishnan2021habitat}-trained Uni-NaVid policy~\citep{zhang2024uni}, and a general-purpose VLM agent (Gemini-3.8-flash~\citep{googledeepmind2026gemini38flash}).
Despite solving the same navigation task, the agents exhibit different weakness profiles, confirming that useful training environments depend on the current policy rather than the task alone.
Across all three agents, weakness-conditioned adaptation consistently outperforms matched random and hard-example alternatives on ProcTHOR.
For Uni-NaVid and Gemini, the gains also extend to HM3D after ProcTHOR adaptation, showing that the resulting improvements are not confined to the environment used for training.
Iterative re-diagnosis shows that the dominant weakness may persist or shift after adaptation, and using the updated diagnosis to guide a second round of environment generation yields further performance gains. Our additional analysis shows that gains are concentrated in weakness-related cases and that rollout calibration focuses training on environments within the policy's learning range.
Together, these results show that policy failures provide not only diagnostic signals, but actionable guidance for constructing training environments that target the policy's remaining weaknesses.

%% file: sections/related_work.tex
\section{Related Work}

\noindent\textbf{Environment Design for Embodied AI.}
Environment design for embodied AI increasingly combines scalable scene construction with controllable, simulation-ready generation~\citep{ye20263d}. Procedural generation and curated synthetic scenes expand environmental diversity and realism~\citep{deitke2022,khanna2024hssd,raistrick2024infinigen,paschalidou2021atiss,tang2024diffuscene}, while language-guided and task-conditioned pipelines automate scene composition and robotic training environments~\citep{feng2023layoutgpt,yang2024holodeck,wang2023robogen,fu2024anyhome}. Physical and interactive constraints further improve scene usability~\citep{yang2024physcene}, and recent agentic pipelines refine generated environments through visual critique, physical validation, and semantic feedback~\citep{yang2026sceneweaver,sun2025layoutvlm,xia2026sage}.
Simulation provides scalable training experience for embodied navigation across a broad range of tasks~\citep{savva2019habitat,shen2021igibson,wijmans2019dd,huang2026phymotion,deitke2022,khanna2024hssd}. Prior work scales this experience through procedurally generated or high-fidelity environments, augmented task configurations and trajectories, synthetic demonstrations, and curriculum-based sampling~\citep{maksymets2021thda,kotar2023entl,ehsani2024spoc,ye2021auxiliary,yadav2022ovrl,morad2021curriculum}. Other efforts improve robustness and transfer through controlled perturbations, domain adaptation, target-conditioned simulation, and synthetic instruction--trajectory data~\citep{chattopadhyay2021robustnav,deitke2020robothor,zhu2019simreal,truong2021bidirectional,deitke2023phone2proc,yang2024holodeck,xu2022environment,wang2023scalevln,wang2025srdf}. These works improve the scalability and controllability of training environments, while our focus is to determine what environments to construct based on the diagnosed weaknesses of a particular pretrained policy.

\noindent\textbf{Failure-Driven Learning and Adaptive Data Generation.} Adaptive curricula and unsupervised environment design use learner feedback to prioritize, generate, and edit training environments~\citep{wang2020enhanced,dennis2020emergent,jiang2021prioritized,parker2022evolving,chung2024adversarial}.
Recent approaches extend this principle through LLM-based environment configuration and code generation~\citep{zala2024envgen,liang2024environment,faldor2025omni}, performance-adaptive navigation curricula~\citep{kang2026simworld}, and trajectory-informed scene editing~\citep{yeo2026towards}.
Related work targets grounding errors through counterfactual scene edits~\citep{zhang2026error} or synthesizes and validates environment wrappers from behavioral diagnoses~\citep{huang2026envharness}.
Our approach extends adaptive environment generation from predefined task-level adaptation to policy-specific failure diagnosis in 3D indoor navigation: rather than assuming known skills or weakness categories, \method{} identifies measurable environmental conditions associated with failures within the same task and validates them as policy-specific weaknesses.
These diagnosed weaknesses determine which environmental factors to modify and which regimes to target during 3D environment construction.
Policy rollouts then calibrate the generated candidates, retaining environments that expose the diagnosed weakness while remaining learnable for the current policy.

%% file: sections/method.tex
\section{Method}
We study how to construct synthetic training environments for a pretrained embodied navigation policy. We begin by formalizing the problem setup (\Cref{sec:problem_setup}). Given a policy $\pi$, our framework identifies environmental features that systematically characterize its failures (\Cref{sec:weakness_discovery}), validates these features as reliable and complementary weakness signals, and uses them to guide the construction of targeted training environments (\Cref{sec:env_generation}). We then collect successful trajectories from the selected environments and use them for policy adaptation (\Cref{sec:policy_adaptation}). Overall, our framework closes the loop from policy behavior to environment design and back to policy learning.

\subsection{Problem Setup}
\label{sec:problem_setup}
We study environment generation for the ObjectNav task~\citep{deitke2022}.
Given a target object category $g$, an agent must navigate from an initial state
$s_0$ through an environment $e$ to locate the target.
We consider a pretrained ObjectNav policy $\pi$ and a controllable procedural
simulator used as the environment-generation space.
The simulator provides a base
distribution $p_{\mathrm{sim}}(e)$ over environments, where each environment
specifies the scene layout, object arrangement, and other task-relevant properties.
An episode is defined as $x=(e,s_0,g)$.
Executing $\pi$ on $x$ produces a rollout
$\tau=(s_0,a_0,s_1,a_1,\ldots,s_T)$
and an outcome $y\in\{0,1\}$, where $y=1$ denotes task success.

\subsection{LLM-Guided Discovery of Executable Weaknesses}
\label{sec:weakness_discovery}
\begin{figure}
    \centering
\includegraphics[width=\linewidth]{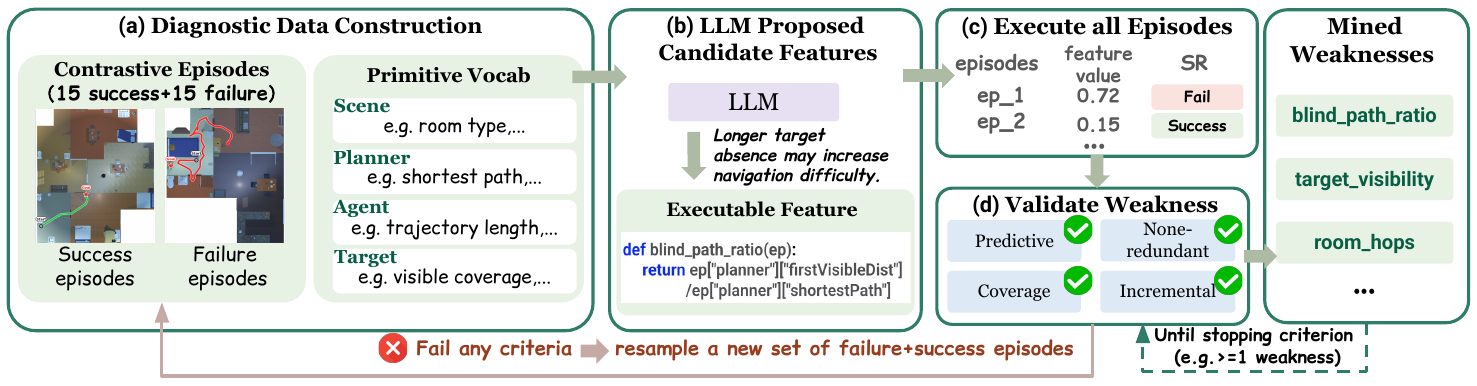}
    \caption{\textbf{LLM-guided discovery and validation of executable policy weaknesses.}
(a) We construct a diagnostic set by contrasting successful and failed episodes
represented with a shared vocabulary of measurable primitives.
(b) An LLM uses these structured examples to propose candidate weakness features
and their executable definitions.
(c) Each candidate is evaluated on the full diagnostic set, and
(d) retained only if it satisfies the validation criteria for coverage,
non-redundancy, predictiveness, and incremental value.
Candidates that fail any criterion trigger a new round of episode sampling and
feature proposal.
This proposal–validation process repeats until the stopping criterion is met, with validated weaknesses retained for subsequent environment generation. 
}
    \label{fig:weakness_discovery}
\end{figure}

Our first step is to identify environment-level factors that systematically
characterize where a pretrained policy struggles.
As illustrated in Fig.~\ref{fig:weakness_discovery}, we use an LLM to reason
over contrasting successful and failed episodes and propose candidate weakness
features, which are then empirically validated on the full diagnostic set.
Candidate weaknesses that fail validation trigger another round of feature proposal using
newly sampled contrastive episodes.
This iterative discovery-and-validation process yields executable weakness
signals that can subsequently guide targeted environment generation.

\noindent\textbf{Diagnostic data construction.}
As shown in \Cref{fig:weakness_discovery}(a), we execute the frozen policy $\pi$ on a held-out set of ObjectNav episodes and
construct a diagnostic dataset $\mathcal{D}=\{(x_i,\tau_i,y_i)\}_{i=1}^{N}$, where $x_i$ denotes an episode, $\tau_i$ is the rollout produced by $\pi$, and $y_i\in\{0,1\}$ is its success outcome.
At each discovery round, we sample 15 successful and 15 failed episodes from $\mathcal{D}$ to form a balanced contrastive set.
To make these episodes directly comparable, we represent each of them using a shared vocabulary of measurable primitives.  
These primitives provide a structured representation of each episode, from which the LLM can define candidate weakness features by selecting and combining relevant measurements.
We organize the primitives into four groups: \emph{scene}, \emph{target}, \emph{path}, and \emph{agent}.
Scene primitives characterize the overall house structure, such as the number of rooms and doors and the start-room type.
Target primitives describe the goal object and its surrounding room, such as the target category, target-room area, and target visibility.
Path primitives characterize the ground-truth shortest path, including its length, the number of room transitions and doors crossed, and the path length before the target first becomes visible.
Finally, agent primitives summarize rollout behavior, such as environment coverage, revisitation, and whether the target room or target was ever reached.
The full primitive vocabulary is provided in Appendix~\Cref{tab:primitives}.

\noindent\textbf{LLM candidate feature proposal.}
Given the sampled episodes and their primitive representations, we prompt an LLM to identify measurable factors that may explain why the policy succeeds in some episodes but fails in others.
As shown in \Cref{fig:weakness_discovery}(b), for each hypothesis, the LLM proposes a candidate weakness feature together with a short rationale and an executable Python function that computes the feature from the available primitives.
The feature may directly use a primitive or combine multiple primitives into a higher-level quantity.
For example, the LLM can define \texttt{blind\_path\_ratio} by combining
\texttt{first\_visible\_path\_length} and \texttt{shortest\_path\_length},
taking their ratio to measure how much of the route must be traversed without visual access to the target. The full prompt for weakness proposal is in Appendix~\ref{app:weakness_prompt}.

\noindent\textbf{Candidate validation and iterative refinement.}
\label{sec:validation}
We execute each proposed feature on the full diagnostic dataset~(\Cref{fig:weakness_discovery}(c)) and retain it only if it provides a reliable and complementary signal of policy failure~(\Cref{fig:weakness_discovery}(d)).
Specifically, candidates are evaluated sequentially for \emph{coverage}, \emph{non-redundancy}, \emph{predictiveness}, and
\emph{incremental value}.
Coverage requires the feature to be computable for most episodes and to vary
across the diagnostic set, while non-redundancy removes features that largely duplicate previously retained weaknesses. We provide the full computation and thresholds for all validation criteria in
Appendix~\ref{sec:weakness_validation}.
For predictiveness, we compare policy success between episodes with low and high
feature values and require a sufficiently large success-rate gap:
\begin{equation}
    \Delta_f =
    \left|
    \mathrm{SR}(\mathcal{L}_f) -
    \mathrm{SR}(\mathcal{H}_f)
    \right|
    \geq \tau_{\Delta},
\end{equation}
where $\mathcal{L}_f$ and $\mathcal{H}_f$ denote the bottom and top terciles of
$f$, respectively.
Finally, incremental value requires the candidate to improve success prediction
beyond the set of already retained features $\mathcal{F}$.
If a candidate fails any validation criterion, we discard it, resample a new set of successful and failed episodes, and query the LLM for another candidate feature. 
We refer to each candidate feature that passes all validation criteria as a
mined weakness, and denote the resulting set for policy $\pi$ as
$\mathcal{F}_{\pi}=\{f_1,\ldots,f_{K_\pi}\}$, where $K_\pi$ is the number of validated weakness features.
These mined weaknesses are used to guide the subsequent environment generation
stage.

\begin{figure}
    \centering
    \includegraphics[width=\linewidth]{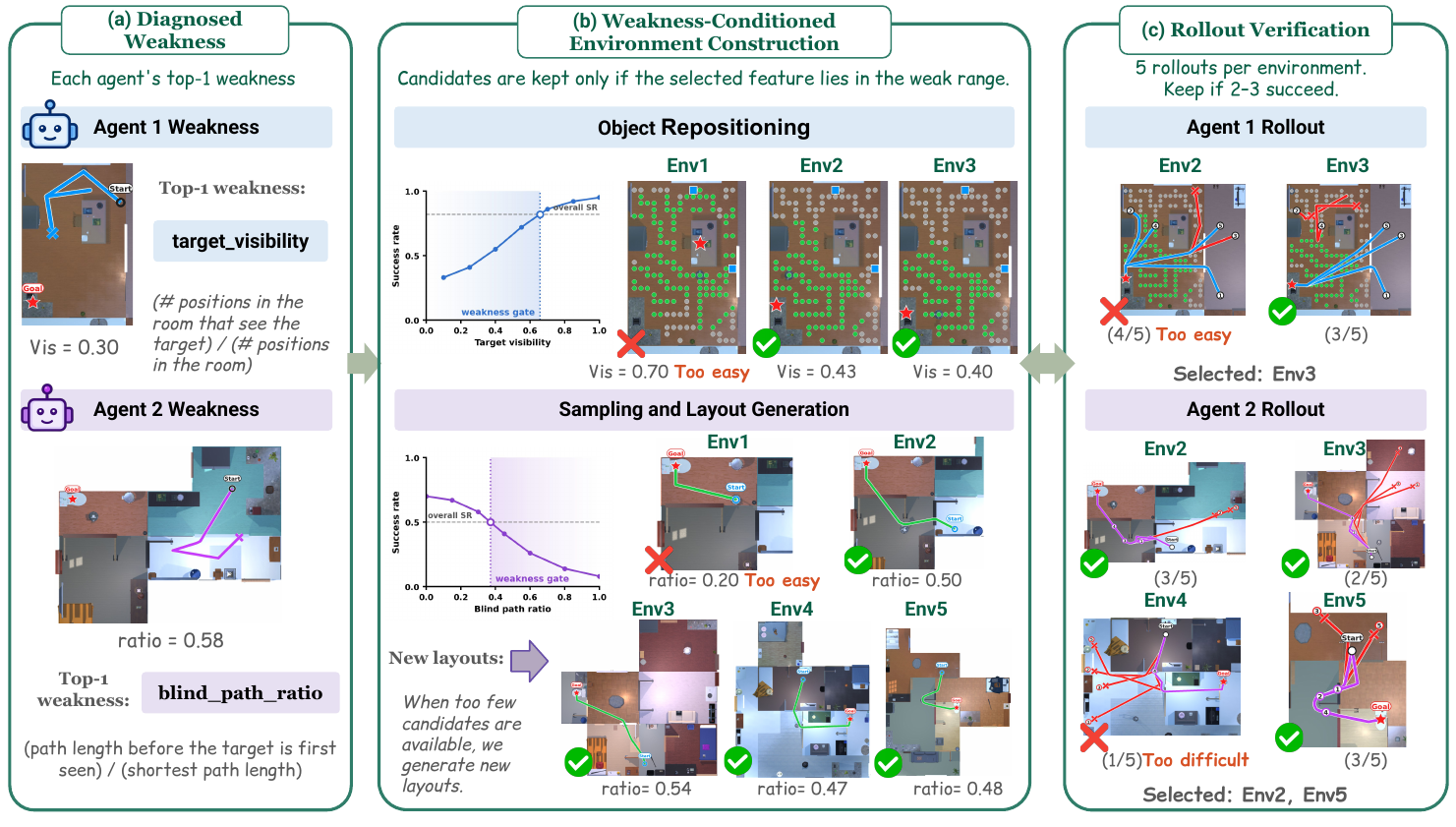}
    \caption{\textbf{Weakness-conditioned environment construction and rollout verification.}
(a) Each policy is associated with its top mined weakness.
(b) Candidate episodes are constructed by modifying the factors relevant to that weakness, such as target placement for \texttt{target\_visibility} or start state and layout for \texttt{blind\_path\_ratio}. Candidates are retained only when their feature values fall within the weakness range, with new layouts generated when existing ones are insufficient.
(c) The policy is rolled out five times in each retained candidate, and episodes with two or three successful rollouts are selected for training.
}
    \label{fig:env_generation}
\vspace{-3mm}
\end{figure}

\subsection{Weakness-Conditioned Environment Generation}
\label{sec:env_generation}

Discovered weaknesses alone do not yet specify useful training data, since
candidate environments may fail to sufficiently express that weakness or may be
too difficult for the current policy to learn from.
We therefore use the selected weakness to determine which environment factors to
modify and retain candidates whose feature values fall within the failure-associated range identified on the diagnostic set. 
We then evaluate the current policy in these environments and keep those that
remain challenging but still solvable.
When suitable candidates are scarce, we generate additional environments to
better cover the desired weakness regime.

\noindent\textbf{Weakness-conditioned environment construction.}
As shown in~\Cref{fig:env_generation}(b), for each policy, we select the mined weakness $f$ with the largest success-rate gap and use it to guide the construction of candidate episodes.
We focus on one weakness per adaptation round because different weaknesses may correspond to distinct environmental factors, and targeting them jointly would make it difficult to isolate which factor drives the observed adaptation gains.
To construct environments that exhibit the selected weakness, we modify the episode factors on which the corresponding feature depends. 
We use three reusable operators for this purpose: re-placing the target object,
sampling a new start state, and generating a new layout.
The applicable operator is determined by which episode factors the weakness
feature depends on.
For example, for Agent 1 in~\Cref{fig:env_generation}, we control
\texttt{target\_visibility} by moving the target within its room to produce
different visibility levels.
For Agent 2, we control \texttt{blind\_path\_ratio} by sampling a new start state or layout, which changes the fraction of the path traveled before the target first becomes visible.

Each operator produces candidate episodes $x'$. We first discard invalid
configurations, such as target placements that are not supported by the scene
or episodes for which no valid navigation path exists.
For the selected weakness $f$, we use the diagnostic set to identify a weak
regime $\mathcal{B}_f$, consisting of feature-value ranges for which the policy's
success rate is lower than its overall success rate on the diagnostic set.
We retain a candidate episode only if
\begin{equation}
    f(x') \in \mathcal{B}_f .
\end{equation}
When the existing layouts provide too few candidates satisfying this constraint,
we generate additional layouts and apply the same criterion.

\noindent\textbf{Rollout verification.}
Satisfying the weakness constraint does not guarantee that a candidate provides a useful training signal, since episodes with similar feature values can still differ substantially in difficulty for the current policy.
As shown in~\Cref{fig:env_generation}(c),  we therefore execute the frozen policy five times in each retained candidate
and count the number of successful rollouts.
Episodes that the policy almost always solves provide little additional training
signal, while episodes that it almost never solves rarely yield successful
experience for adaptation.
We therefore retain episodes with intermediate empirical success:
$2 \;\le\; n_{\mathrm{succ}}(x') \;\le\; 3,
$
where $n_{\mathrm{succ}}(x')$ denotes the number of successful rollouts among
the five trials. These rollouts are used only for episode selection.

\subsection{Training Data Construction and Policy Adaptation}
\label{sec:policy_adaptation}

The selected episodes from~\Cref{sec:env_generation} determine what experience each policy learns from, while we preserve its original adaptation mechanism.
This keeps the framework independent of a particular learning algorithm and allows our method and controls to differ only in how training episodes are constructed.
For reinforcement learning policies, we mix the selected episodes with the
original training distribution and continue training with the original RL objective. 
For imitation-based policies, we fine-tune on successful self-rollouts collected from the selected episodes. 
For models without parameter access, such as closed-weight vision and language models, we use successful trajectories from the selected episodes as in-context
examples instead of weight updates.
After adaptation, the updated policy can be evaluated again and passed through the same weakness discovery procedure to identify remaining weaknesses.

%% file: sections/experiment.tex
\section{Experiment}

\subsection{Experimental Setup}
\label{sec:exp_setup}

\noindent\textbf{Agents.} We evaluate our framework on three different navigation agents: a ProcTHOR ObjectNav policy trained with DD-PPO~\citep{zeng2024poliformer}, the HM3D-trained navigation policy Uni-NaVid~\citep{zhang2024uni}, and the general-purpose vision-language model Gemini-3.8-flash~\citep{googledeepmind2026gemini38flash}.
(1) The ProcTHOR policy represents an in-domain reinforcement-learning agent, allowing us to test whether our framework can identify weaknesses that remain after in-domain training.
(2) Uni-NaVid is trained on HM3D but diagnosed and adapted in ProcTHOR, enabling us to study whether weakness-targeted synthetic training transfers back to its original evaluation domain.
(3) Finally, Gemini serves as a general-purpose navigation agent, for which adaptation is performed via in-context demonstrations rather than parameter updates.
Together, these agents allow us to examine whether the same weakness-discovery framework can identify distinct failure profiles across heterogeneous policies and support different forms of adaptation.

\input{tables/mainresult}

\noindent\textbf{Experimental protocol and implementation details.}
We use ProcTHOR for weakness diagnosis and training environment construction because its procedural structure allows us to directly modify the environmental factors underlying the diagnosed weaknesses.
For each agent, we execute the frozen policy on a held-out diagnostic set and use Gemini-2.5-Pro~\citep{comanici2025gemini} to propose executable weakness features.
(1) For DD-PPO, we continue reinforcement learning from the pretrained checkpoint for 10M environment steps.
(2) For Uni-NaVid, we restrict ProcTHOR training episodes to the six object categories shared with HM3D and fine-tune the model with LoRA on its own successful trajectories.
(3) For Gemini, successful weakness-matched trajectories are converted into in-context examples and appended to the system prompt. See~\Cref{app:gemini} for more details. 
We compare against matched baselines under the same adaptation budget and report Success Rate (SR) and Success weighted by Path Length (SPL)~\citep{zeng2024poliformer}.
Specifically, WEDGE uses 668 calibrated environments for DD-PPO, 2,480 successful trajectories for Uni-NaVid, and one worked example (selected from 64 successful practice trajectories) for Gemini. All baselines use the same number of training environments, trajectories, or in-context examples.
We compare against random and hard-mined baselines. The hard-mined baselines choose training data based only on whether the agent succeeds or fails, without using any weakness features. For DD-PPO, we pick the environments where the agent succeeds least often. For Uni-NaVid and Gemini, we take episodes the agent previously failed, run it again, and keep the successful trajectories.
Evaluation uses 1,550 ProcTHOR episodes for DD-PPO, 655 ProcTHOR and 1,000 HM3D
episodes for Uni-NaVid, and 300 ProcTHOR and 300 HM3D episodes for Gemini.
Appendix~\ref{extra_experiments} provides detailed implementation.

\subsection{Policy-Specific Weakness Discovery}
\label{sec:exp_weakness}

\Cref{tab:main_results} and \Cref{tab:weaknesses} in the Appendix report the top discovered weakness for each agent together with its corresponding success-rate gap.
For the ProcTHOR-trained DD-PPO policy, the top weakness is \texttt{target\_visibility}, with a 28.5 SR gap.
The discovery procedure also identifies \texttt{doorway\_visibility} as a complementary weakness, indicating that DD-PPO is particularly sensitive to whether the target can be observed from within or near the target room.
For Uni-NaVid, the top weakness is \texttt{blind\_path\_ratio}, which produces a 37.3 SR gap, followed by \texttt{room\_hops} with a 34.8 SR gap.
Gemini also exhibits \texttt{blind\_path\_ratio} as its dominant weakness, with a 37.6 SR gap.
These results show that agents solving the same navigation task can exhibit different weakness profiles, motivating policy-specific environment generation.

\subsection{Weakness-Conditioned Adaptation}
\label{sec:adaptation}

\noindent\textbf{In-domain adaptation.}
We first evaluate whether weakness-conditioned experience improves adaptation
in the ProcTHOR environment used for diagnosis and generation.
As shown in Table~\ref{tab:main_results}, \method{} consistently outperforms random and hard-example alternatives across all three agents.
For the ProcTHOR-trained DD-PPO policy, continued training on random or
hard-mined environments yields little improvement over the pretrained model
(82.1 SR), whereas \method{} improves SR to 85.6 and SPL from 58.2 to 59.8.
Uni-NaVid shows larger gains from weakness-conditioned adaptation.
Because its original HM3D training data are not publicly available, adaptation must rely on ProcTHOR experience, introducing an inherent domain shift.
We therefore use random self-rollout adaptation on ProcTHOR as the matched
baseline, controlling for continued training with the same data source and fine-tuning budget. While random self-rollout reduces SR from 51.6 to 46.6, weakness-conditioned adaptation instead improves SR from 51.6 to 55.7 and SPL from 35.3 to 38.5.
For the closed-weight Gemini agent, weakness-targeted in-context examples
improve SR from 41.0 to 43.7, outperforming both random examples (40.7)
and hard examples (41.8).
Together, these results show that the benefit comes from targeting the policy's diagnosed weaknesses rather than simply providing additional or difficult adaptation data.
We further ablate mined versus generated environments in Appendix~\Cref{tab:mined_generated}, showing that mining existing episodes alone is insufficient and that generating new environments targeted to the diagnosed weakness provides additional benefit.

\noindent\textbf{Evaluation on HM3D after ProcTHOR adaptation.}
We further evaluate whether the gains from ProcTHOR adaptation extend to HM3D without extra training.
For Uni-NaVid, this comparison reveals an important difference between hard-example mining and weakness-conditioned adaptation.
Although hard-example mining improves performance on ProcTHOR, the improvement does not extend to HM3D, where SR remains at 69.0 compared with 69.2 for the pretrained policy.
Random self-rollout performs even worse, reducing HM3D SR to 67.1.
In contrast, \method{} improves performance on both domains, reaching 73.4 SR on HM3D while maintaining comparable SPL to the pretrained policy (28.6 vs.\ 28.5).
Gemini shows a similar pattern: random and hard examples achieve 58.0 and 60.9 SR on HM3D, respectively, whereas weakness-conditioned examples improve SR to 63.5 and SPL to 29.1.
These results suggest that weakness-conditioned adaptation captures training signals that remain useful beyond the ProcTHOR environments used for adaptation.

\begin{figure*}[t]
\centering
\begin{minipage}[t]{0.58\textwidth}
    \centering
    \includegraphics[width=\linewidth]{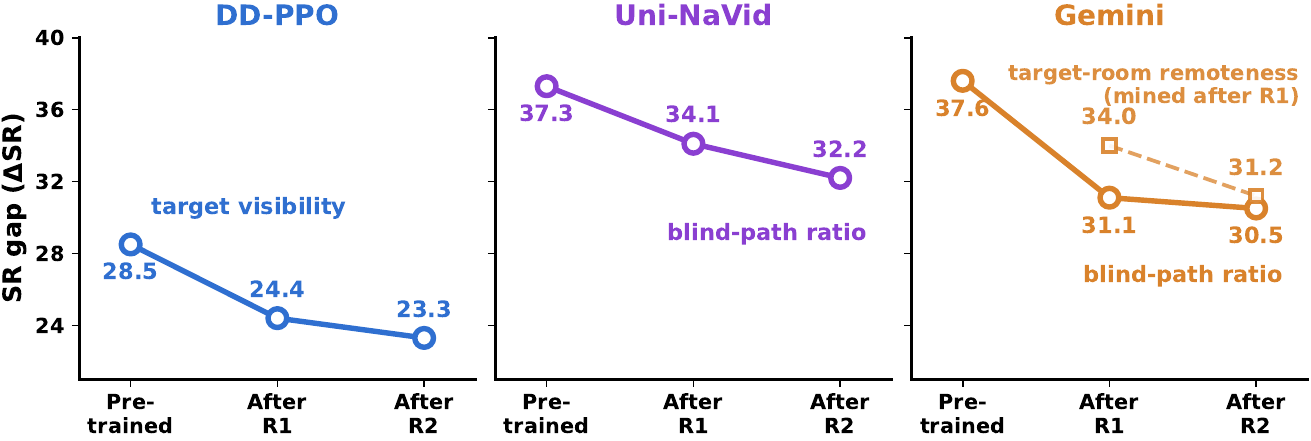}
    \vspace{-2pt}
    (a) Weakness evolution across adaptation rounds
\end{minipage}
\hfill
\begin{minipage}[t]{0.4\textwidth}
    \centering
    \includegraphics[width=\linewidth]{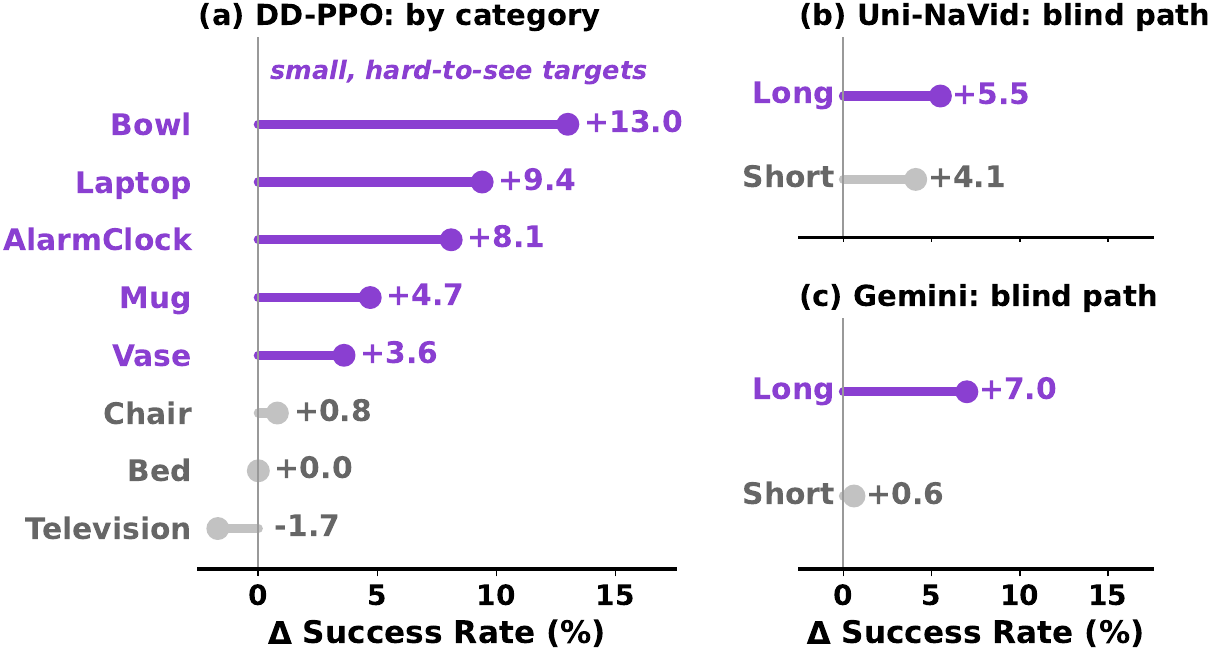}
    \vspace{-2pt} 
  (b) Gains under diagnosed weaknesses
\end{minipage}
\vspace{4pt}
\caption{
\textbf{Analysis of weakness-conditioned adaptation.}
\textbf{(a)} We re-diagnose each agent after every adaptation round and track
the SR gap associated with its dominant weakness. DD-PPO and Uni-NaVid retain
the same dominant weakness, while Gemini develops a new weakness after the
first round.
\textbf{(b)} Adaptation gains are larger in weakness-associated regimes.
For Uni-NaVid and Gemini, we compare gains across
\texttt{blind\_path\_ratio} regimes. For DD-PPO, gains are reported by target
category, with larger improvements on small, hard-to-see targets.
}
\label{fig:weakness_analysis}

\end{figure*}

\noindent\textbf{Iterative weakness re-diagnosis.}
Finally, we examine how the diagnosed weaknesses evolve across adaptation rounds.
As shown in~\Cref{fig:weakness_analysis}(a), for DD-PPO, the top weakness remains \texttt{target\_visibility} after the first round, while both SR and SPL effectively saturate after the first round of adaptation.
Uni-NaVid also retains \texttt{blind\_path\_ratio} as its top weakness across both rounds,
while continuing to improve from 54.5 to 55.7 SR on ProcTHOR.
Gemini exhibits a change in its weakness profile:
\texttt{blind\_path\_ratio} is the dominant weakness in the first round, whereas
\texttt{target\_room\_remoteness} becomes the top weakness after adaptation.
These results show that iterative re-diagnosis can either confirm a remaining
weakness or reveal a new limiting factor as the policy improves, providing the
target for the next round of environment generation.

\subsection{Analysis and Ablations}

\noindent\textbf{Gains under diagnosed weakness.}
We further examine how performance changes across regimes associated with each agent's diagnosed weakness. As shown in~\Cref{fig:weakness_analysis}(b),
the gains are especially large in these challenging regimes.
For DD-PPO, improvements are largest on small, hard-to-see targets, including Bowl, Laptop, and AlarmClock, while several larger objects change little.
The trend is especially clear for Gemini, where long-blind-path episodes improve by +7.0 SR, compared with little or no gain on shorter paths.
These results show that weakness-conditioned adaptation not only improves aggregate performance, but also produces substantial gains on the failure regimes targeted during adaptation.

\noindent\textbf{Effect of rollout-based calibration.}
As shown in~\Cref{tab:calibration}, weakness matching alone does not ensure that an environment provides a useful
learning signal. Many weakness-matched environments are either already easy
for the current policy or remain beyond its current capability.
Rollout-based calibration filters these extremes and focuses adaptation on
environments of intermediate difficulty.
For DD-PPO, the calibrated subset contains only 16\% of the weakness-matched environments, yet achieves higher performance than using the full set (85.4 vs. 84.4 SR).
We observe the complementary effect for Uni-NaVid, where increasing the share
of the hardest episodes does not improve performance and instead slightly
degrades overall results.
Together, these results suggest that effective adaptation depends not only on targeting a policy weakness, but also on matching the generated environments to
the policy's current learning range.

\begin{wraptable}{r}{0.53\textwidth}
\vspace{-16pt}
\centering
\footnotesize
\setlength{\tabcolsep}{3pt}
\renewcommand{\arraystretch}{1.0}

\caption{
\textbf{Effect of rollout calibration.}
Proc. and HM3D report SR; Hardest: share of hardest episodes.
}
\label{tab:calibration}

\begin{tabular*}{\linewidth}{
@{\extracolsep{\fill}}lcc@{}
}
\toprule
\textbf{DD-PPO}
& \textbf{\# Env.}
& \textbf{SR} $\uparrow$ \\
\midrule
Pretrained
& -- & 82.1 \\
All weakness-matched
& 4,215 & 84.4 \\
Calibrated (2--3/5)
& 668 & \textbf{85.4} \\
\bottomrule
\end{tabular*}

\par\vspace{3pt}

\begin{tabular*}{\linewidth}{
@{\extracolsep{\fill}}lccc@{}
}
\toprule
\textbf{Uni-NaVid}
& \textbf{Hardest}
& \textbf{Proc.}
& \textbf{HM3D} \\
\midrule
Random self-rollout
& -- & 46.6 & 67.1 \\
Weakness-matched
& 12.1\% & \textbf{49.0} & \textbf{70.0} \\
Hard-heavy
& 25.1\% & 47.2 & 69.2 \\
\bottomrule
\end{tabular*}
\end{wraptable}

\noindent\textbf{Computational efficiency and generation scale}. Weakness discovery adds little overhead, as it operates on already-collected diagnostic rollouts and requires only a small number of LLM calls. More importantly, it reduces the cost of identifying useful training environments: to obtain the same number of hard target placements, unguided hard-example mining required evaluating roughly twice as many candidates (8.4k vs. 4.2k), corresponding to 42k versus 21k policy rollouts. 
We provide a full breakdown of generation, rollout, and adaptation costs in Appendix~\Cref{app:computation_cost}.

%% file: tables/mainresult.tex
\begin{table*}[t]
\centering
\small
\setlength{\tabcolsep}{4pt}
\renewcommand{\arraystretch}{1.10}

\caption{
\textbf{Experimental results on policy-specific adaptation and generalization.}
For each agent, we report the validated weaknesses identified in ProcTHOR, adaptation performance on ProcTHOR, and performance when evaluated back on HM3D. \textbf{Bold} denotes the best performance.
$^{\dagger}$ Uni-NaVid's original HM3D training data are unavailable, so the pretrained model is shown as a reference.
}
\label{tab:main_results}

\resizebox{0.85\textwidth}{!}{%
\begin{tabular}{
>{\raggedright\arraybackslash}p{1.8cm}
>{\centering\arraybackslash}p{3.0cm}
>{\raggedright\arraybackslash}p{3.0cm}
cccc
}
\toprule

\multirow{2}{*}{\textbf{Agent}}
& \multirow{2}{*}{\textbf{Diagnosed weakness}}
& \multirow{2}{*}{\textbf{Adaptation setting}}
& \multicolumn{2}{c}{\textbf{ProcTHOR}}
& \multicolumn{2}{c}{\textbf{HM3D}} \\

\cmidrule(lr){4-5}
\cmidrule(lr){6-7}

& &
& \textbf{SR} $\uparrow$
& \textbf{SPL} $\uparrow$
& \textbf{SR} $\uparrow$
& \textbf{SPL} $\uparrow$ \\

\midrule

\multirow{5}{=}{\textbf{DD-PPO}}
&
\multirow{5}{=}{
\shortstack[c]{
\textbf{Target visibility} \\
($\Delta$SR = 28.5)
}}
&
Pretrained
& 82.1 & 58.2
& -- & -- \\

&
&
Random env.
& 82.3 & 58.8
& -- & -- \\

&
&
Hard-mined
& 82.4 & 58.1
& -- & -- \\

\cmidrule(lr){3-7}
& & \ourscell{\method{} R1 (Ours)}
& \ourscell 85.4 & \ourscell \textbf{59.8}
& \ourscell -- & \ourscell -- \\

& & \ourscell{\method{} R2 (Ours)}
& \ourscell\textbf{85.6} & \ourscell\textbf{59.8}
& \ourscell -- & \ourscell -- \\

\midrule

\multirow{5}{=}{\mbox{\textbf{Uni-NaVid}}}
&
\multirow{5}{=}{
\shortstack[c]{
\textbf{Blind-path ratio}\\
($\Delta$SR = 37.3)
}}
&
Pretrained$^{\dagger}$
& 51.6 & 35.3
& 69.2 & 28.5 \\

&
&
Random self-rollout
& 46.6 & 34.1
& 67.1 & 26.7 \\

&
&
Hard self-rollout
& 53.1 & 36.9
& 69.0 & 26.2 \\

\cmidrule(lr){3-7}
& & \ourscell{\method{} R1 (Ours)}
& \ourscell 54.5 & \ourscell 37.3
& \ourscell 71.4 & \ourscell 28.4 \\

& & \ourscell{\method{} R2 (Ours)}
& \ourscell\textbf{55.7} & \ourscell\textbf{38.5}
& \ourscell\textbf{73.4} & \ourscell\textbf{28.6} \\

\midrule

\multirow{4}{=}{\textbf{Gemini}}
&
\multirow{4}{=}{
\shortstack[c]{
\textbf{Blind-path ratio} \\
($\Delta$SR = 37.6)
}}
&
Zero-shot
& 41.0 & 31.8
& 57.9 & 27.0 \\

&
&
Random example
& 40.7 & 32.6
& 58.0 & 27.9 \\
&
&
Hard example
& 41.8 & 32.7
& 60.9 & 28.7 \\

\cmidrule(lr){3-7}
& & \ourscell{\method{} R1 (Ours)}
& \ourscell43.7 & \ourscell33.4
& \ourscell63.5 & \ourscell{29.1} \\
& & \ourscell{\method{} R2 (Ours)}
& \ourscell\textbf{44.3} & \ourscell\textbf{33.7} & \ourscell\textbf{64.2} & \ourscell\textbf{30.2}
\\

\bottomrule
\end{tabular}%
}
\vspace{-3mm}
\end{table*}

%% file: sections/conclusion.tex
\section{Conclusion}
We introduced \method{}, a weakness-guided framework for generating training environments for navigation policies. By diagnosing policy-specific failure conditions and using them to guide environment construction, \method{} improves adaptation across different agents and learning mechanisms. Our results further show that these gains can transfer across simulators and that re-diagnosis can support continued adaptation over multiple rounds.
More broadly, our findings suggest that agent failures can be used as actionable signals for targeted improvement, enabling training environments to be constructed around what each policy still needs to learn.

\section{Acknowledgement}
This work was supported by NSF-AI Engage Institute DRL-2112635, NSF-CAREER Award 1846185, ARO Award W911NF2110220, ONR Grant N00014-23-1-2356, and NDSEG PhD Fellowship (Air Force Office of Scientific Research under award number FA9550-25-C-B010). The views contained in this article are those of the authors and not of the funding agency.

%% file: sections/appendix.tex
\section{Appendix}

\subsection{Weakness Feature Validation}
\label{sec:weakness_validation}

Given a candidate feature $f$ proposed by the LLM, we evaluate it over the full
diagnostic dataset $\mathcal{D}=\{(x_i,y_i)\}_{i=1}^{N}$ before accepting it as
a policy weakness. The validation procedure consists of four sequential criteria:
coverage, non-redundancy, predictiveness, and incremental value. A candidate is
rejected as soon as it fails any criterion.

\paragraph{Coverage.}
A candidate feature should be defined for nearly all diagnostic episodes.
Let $\mathcal{D}_f \subseteq \mathcal{D}$ denote the episodes for which
$f$ can be computed. We require
\begin{equation}
    \frac{|\mathcal{D}_f|}{|\mathcal{D}|} \geq \tau_{\mathrm{cov}},
\end{equation}
where $\tau_{\mathrm{cov}}=0.95$.
We additionally discard features that are constant over $\mathcal{D}_f$.

\paragraph{Non-redundancy.}
To avoid retaining multiple features that capture essentially the same
failure pattern, we compare each candidate $f$ with the previously retained
features $\mathcal{F}$. We require
\begin{equation}
    \max_{f' \in \mathcal{F}} |\rho_s(f,f')| < \tau_{\rho},
\end{equation}
where $\rho_s$ is the Spearman correlation computed over the diagnostic set.
We set $\tau_{\rho}=0.85$. The criterion is skipped when $\mathcal{F}$ is empty.

\paragraph{Predictiveness.}
A useful weakness should separate episodes on which the policy performs
substantially differently. We divide episodes into the bottom and top terciles
of $f$, denoted by $\mathcal{L}_f$ and $\mathcal{H}_f$, and measure the
corresponding success-rate gap:
\begin{equation}
    \Delta_f =
    \left|
    \mathrm{SR}(\mathcal{L}_f)
    -
    \mathrm{SR}(\mathcal{H}_f)
    \right|.
\end{equation}
We retain candidates with $\Delta_f \geq \tau_{\Delta}$ and a significant
difference in success rate ($p<0.05$, two-proportion $z$-test).
We additionally check that this relationship generalizes beyond the episodes
used to identify it. We estimate the tercile thresholds on a random 70\% split
of the diagnostic set and apply the same thresholds to the remaining 30\%.
The candidate must preserve the direction of the success-rate difference and
achieve a gap of at least $\tau'_{\Delta}$ on the held-out split.
We use $\tau_{\Delta}=10$ and $\tau'_{\Delta}=5$ percentage points.

\paragraph{Incremental value.}
Finally, a new weakness should provide information beyond the features already
retained in $\mathcal{F}$. We fit a logistic-regression predictor of episode
success using the retained features and evaluate it with five-fold
cross-validated AUC. We accept $f$ only if
\begin{equation}
    \mathrm{AUC}_{\mathrm{cv}}(\mathcal{F}\cup\{f\})
    -
    \mathrm{AUC}_{\mathrm{cv}}(\mathcal{F})
    \geq \tau_{\mathrm{AUC}} .
\end{equation}
We set $\tau_{\mathrm{AUC}}=0.01$ and use $0.5$ as the baseline AUC when
$\mathcal{F}$ is empty.

\begin{table}[t]
\centering
\small
\setlength{\tabcolsep}{8pt}
\renewcommand{\arraystretch}{1.12}
\caption{\textbf{Ablation of mined and generated adaptation on ProcTHOR}}
\label{tab:mined_generated}
\begin{tabular}{@{}lcc@{}}
\toprule
\textbf{Adaptation data} & \textbf{SR} $\uparrow$ & \textbf{SPL} $\uparrow$ \\
\midrule
Random self-rollout & 46.6 & 34.1 \\
Weakness-mined existing   & 49.2  & 34.9  \\
Weakness-generated only    & 51.3  & 35.4 \\
Mined + Generated (\method{}) & 54.5 & 37.3 \\
\bottomrule
\end{tabular}
\end{table}

\subsection{Extra Experiments}
\label{extra_experiments}

\subsubsection{Computational cost.}
\label{app:computation_cost}

The additional overhead introduced by weakness discovery is small relative to environment evaluation and policy adaptation. Discovery operates on already-logged diagnostic episodes, runs on CPU, and requires only about ten LLM calls per agent. The dominant cost instead comes from rollout-based candidate verification and subsequent adaptation. Importantly, weakness-guided filtering reduces this downstream search cost: to obtain the same number of hard target placements, unguided hard-example mining required evaluating roughly twice as many candidates (8.4k vs. 4.2k candidates; 42k vs. 21k rollouts). Across agents, one complete adaptation round requires approximately one GPU-day for DD-PPO, two GPU-days for Uni-NaVid, and half a day plus about \$110 in API cost for Gemini.

For DD-PPO, weakness-conditioned generation produces 4,215 target placements across 2,400 ProcTHOR houses. For Uni-NaVid, we first mine a large pool of 296,823 existing start--target pairs and further augment this experience with 7,147 episodes from 2,716 newly generated houses. After rollout and quality filtering, the existing and generated pools contribute 1,630 and 850 successful trajectories, respectively, yielding 51,162 training samples in total. For Gemini, 400 weakness-matched practice episodes yield 64 successful trajectories, from which one is selected as the worked example for in-context adaptation.

\subsubsection{Gemini Navigation Agent}
\label{app:gemini}

\paragraph{VLM Navigator.}
We use Gemini-3.8-flash with low reasoning effort.
At each decision step in ProcTHOR, the simulator captures four RGB views
separated by $90^\circ$ (front, right, back, and left), with resolution of 576*768.
The model receives these views together with its previous 16 decisions and predicts whether the target is visible and one action from \{\texttt{FORWARD}, \texttt{RIGHT}, \texttt{BACK}, \texttt{LEFT},
\texttt{END}\}.
A directional action first orients the agent toward the selected direction and
then executes up to three 0.25\,m forward steps, stopping early when blocked.
Episodes are capped at 60 decisions.
The agent may predict \texttt{END} only when it also predicts that the target is visible in the front view.
Before accepting \texttt{END}, we make a second model call using the current
front view and an additional $30^\circ$ downward-facing view.
The model must confirm that the target is clearly identifiable and within approximately 1\,m.
A rejected stop consumes one decision and is recorded in the action history.

\paragraph{In-context adaptation.}
For weakness-conditioned adaptation, Gemini interacts once with 400 weakness-matched ProcTHOR practice episodes, of which 64 are successful.
From the successful practice episodes, we select one representative trajectory
and compress its decision sequence into a worked example.
Gemini-2.5-Pro additionally produces a two-sentence explanation of why the
trajectory succeeds.
The resulting worked example is appended to the navigation prompt and is used
for the \method{} results reported in the main table.
As an alternative adaptation format, we also derive three navigation rules and
a target-specific search prior from successful practice episodes.

\paragraph{Iterative in-context adaptation.}
For Gemini, each adaptation round adds one weakness-conditioned worked example to the navigation prompt. In Round 1, we run the base agent on the weakness-matched practice set and select one successful trajectory, which is compressed into a decision sequence and paired with a short explanation generated by Gemini-2.5-Pro. After adapting the agent with this example, we re-diagnose its remaining weakness and collect a new weakness-matched practice set using the updated Round 1 agent. We then select a second successful trajectory from this new set and append it as an additional worked example. Thus, the Round 2 agent receives both the Round 1 and Round 2 examples in its prompt.

\begin{tcolorbox}[
    colback=gray!5,
    colframe=gray!50,
    boxrule=0.6pt,
    arc=2pt,
    left=6pt,
    right=6pt,
    top=6pt,
    bottom=6pt,
    title=\textbf{Gemini Navigation Prompt}
]
\small
\ttfamily

You are a home robot performing ObjectNav. Your goal: find a \{target\} and stop right in front of it.

You are given FOUR views taken from where you stand: FRONT, RIGHT, BACK, LEFT (90 degrees apart).

Recent moves: \{history\}

\medskip
Answer in EXACTLY this format (two lines):

VISIBLE: <yes or no>

ACTION: <FORWARD, RIGHT, BACK, LEFT, or END>

\medskip
Rules:

- Choose END if and only if VISIBLE is yes.

- If you can see the \{target\} in ANY view but it is far, move toward it instead of ending.

- Otherwise EXPLORE: pick the direction with the most unexplored space.

- If recent moves show a loop, commit to leaving through an unexplored doorway.

- Avoid directions marked as blocked.

\end{tcolorbox}

\begin{tcolorbox}[
    colback=gray!5,
    colframe=gray!50,
    boxrule=0.6pt,
    arc=2pt,
    title=\textbf{Round 1 Worked Example}
]
\small
\ttfamily

WORKED EXAMPLE --- one of your own successful episodes.
The target was invisible for the first half of the route.

Decisions:
FORWARD FORWARD FORWARD | RIGHT |
FORWARD FORWARD FORWARD | LEFT |
FORWARD (hit wall) | RIGHT | (scan) |
FORWARD FORWARD |
END-rejected (too far) |
FORWARD | END -> SUCCESS.

Why it worked:
Long committed FORWARD runs through openings covered two rooms quickly.
After the wall hit, the agent turned once and kept moving instead of oscillating.
After the rejected END, it stepped closer and only then re-declared END.

\end{tcolorbox}

\begin{tcolorbox}[
    colback=gray!5,
    colframe=gray!50,
    boxrule=0.6pt,
    arc=2pt,
    title=\textbf{Round 2 Worked Example}
]
\small
\ttfamily

WORKED EXAMPLE 2 --- another of your own successful episodes.
The target was located in a relatively small room within a much larger house, requiring sustained exploration to find.

Decisions:
LEFT | RIGHT | LEFT | BACK (blocked) x2 |
LEFT | RIGHT | FORWARD x3 | BACK |
LEFT | FORWARD x3 | RIGHT | LEFT | RIGHT |
END -> SUCCESS.

Why it worked:
When the path was blocked, the agent changed direction and continued exploring instead of repeatedly searching the same nearby area. It kept progressing through unexplored spaces until reaching the remote target room.

\end{tcolorbox}

\begin{table*}[t]
\centering
\small
\setlength{\tabcolsep}{6pt}
\renewcommand{\arraystretch}{1.15}

\caption{
\textbf{Policy-specific weaknesses.}
Validated weakness features and the corresponding success-rate drop between
the diagnosed weak regime and the contrasting regime.
\textbf{Bold} denotes the top-ranked weakness used for environment generation.
}
\label{tab:weaknesses}

\resizebox{\textwidth}{!}{
\begin{tabular}{@{}llp{5.50cm}c@{}}
\toprule
\textbf{Agent} &
\textbf{Discovered weakness} &
\textbf{Executable definition} &
\textbf{SR drop} \\
\midrule

\multirow{2}{*}{
\shortstack[l]{DD-PPO}
}
&
\textbf{\texttt{target\_visibility}}
&
Fraction of standing positions in the target room from which the target is visible
&
\textcolor{red!70!black}{\textbf{$\downarrow 28.5$}}
\\

&
\texttt{doorway\_visibility}
&
Whether the target is visible from a doorway of the target room, evaluated within low-visibility rooms
&
\textcolor{red!70!black}{$\downarrow 22.5$}
\\

\midrule

\multirow{2}{*}{
\shortstack[l]{Uni-NaVid}
}
&
\textbf{\texttt{blind\_path\_ratio}}
&
Fraction of the shortest path traversed before the target first becomes visible
&
\textcolor{red!70!black}{\textbf{$\downarrow 37.3$}}
\\

&
\texttt{room\_hops}
&
Number of room-to-room transitions along the shortest path to the target
&
\textcolor{red!70!black}{$\downarrow 34.8$}
\\

\midrule

\shortstack[l]{Gemini}
&
\textbf{\texttt{blind\_path\_ratio}}
&
Fraction of the shortest path traversed before the target first becomes visible
&
\textcolor{red!70!black}{\textbf{$\downarrow 37.6$}}
\\

\bottomrule
\end{tabular}
}

\vspace{2pt}
\end{table*}

\begin{table}[t]
\centering
\small
\setlength{\tabcolsep}{4pt}
\renewcommand{\arraystretch}{1.15}
\caption{\textbf{Measurement primitives.} Every evaluated episode is summarised by a fixed
vocabulary of simulator-measured quantities. \emph{Scene}, \emph{Target} and \emph{Path}
primitives depend only on the layout and the ground-truth shortest path, so they are
controllable at generation. }
\label{tab:primitives}
\begin{tabular}{@{}p{0.12\linewidth} p{0.40\linewidth} p{0.45\linewidth}@{}}
\toprule
\textbf{Group} & \textbf{Primitive} & \textbf{Meaning} \\
\midrule
\multirow{6}{=}{\textbf{Scene}\\\footnotesize house structure}
 & \texttt{houseArea} & floor area of the house (m$^2$) \\
 & \texttt{roomsTotal} & number of rooms \\
 & \texttt{numDoorsHouse} & number of doors in the house \\
 & \texttt{startRoomType} & room type the episode starts in \\
 & \texttt{roomPair} & start room type $\rightarrow$ target room type \\
 & \texttt{numSameTypeRooms} & rooms of the same type as the target room \\
\midrule
\multirow{10}{=}{\textbf{Target}\\\footnotesize object and its room}
 & \texttt{targetCategory} & goal object category \\
 & \texttt{numTargetObjs} & instances of that category in the house \\
 & \texttt{numPlausibleDecoyRooms} & rooms that plausibly contain the category but do not \\
 & \texttt{targetRoomArea} & area of the target's room (m$^2$) \\
 & \texttt{targetRoomDoors} & doors of the target's room \\
 & \texttt{targetRoomIsLeaf} & target room is a dead end \\
 & \texttt{targetInCorner} & target sits in a room corner \\
 & \texttt{targetDistToNearestDoor} & distance from target to the nearest door (m) \\
 & \texttt{visibilityCoverage} & fraction of standing spots in the room with a clear line to the target \\
 & \texttt{occlusion}, \texttt{occlusionDegree$^{\ast}$}, \texttt{losOnsetRange} & how much the target is blocked, and the range at which it first becomes visible \\
 &\texttt{targetRoomRemoteness} & how much of the house lies outside the target's room, relative to the room's own area (a small room in a large house scores high) \\
\midrule
\multirow{8}{=}{\textbf{Path}\\\footnotesize ground-truth shortest path}
 & \texttt{shortestPath} & geodesic start-to-target distance (m) \\
 & \texttt{roomHopsStartToTarget} & rooms traversed on the path \\
 & \texttt{doorsOnPath} & doors crossed on the path \\
 & \texttt{branchingBeforeTarget} & junctions passed before reaching the target \\
 & \texttt{decoyBranchesBeforeTarget} & junctions whose other branch leads to a decoy room \\
 & \texttt{doorwaySeesTarget} & target is visible from a doorway of its room \\
 & \texttt{doorwaySeeingFrac} & fraction of the room's doorways that see the target \\
 & \texttt{pathBeforeFirstVisible} & path length walked before the target first becomes visible (m) \\
\midrule
\multirow{8}{=}{\textbf{Agent}\\\footnotesize \emph{context only}}
 & \texttt{envSteps}, \texttt{numActions} & episode length \\
 & \texttt{minDistToTarget} & closest approach to the target (m) \\
 & \texttt{roomsVisited}, \texttt{houseCoverage} & how much of the house was explored \\
 & \texttt{revisitRatio}, \texttt{stuckFrac}, \texttt{fullSpins} & looping, getting stuck, full in-place spins \\
 & \texttt{wrongRoomEntries}, \texttt{stayedInStartRoomSteps} & search behaviour \\
 & \texttt{everReachedTargetDoorway}, \texttt{reachedTargetRoom} & whether the target's room was reached \\
 & \texttt{everVisible}, \texttt{calledEnd} & target ever on screen; agent called stop \\
 & \texttt{forwardShare}, \texttt{turnShare} & action mix \\
\bottomrule
\end{tabular}
\\[2pt]
\footnotesize $^{\ast}$\texttt{occlusionDegree} is logged as mean and max over standing spots.
\end{table}

\subsection{Prompt for Weakness Feature Proposal}
\label{app:weakness_prompt}

\begin{tcolorbox}[
    colback=gray!4,
    colframe=gray!50,
    boxrule=0.6pt,
    arc=2pt,
    title=\textbf{Prompt for Weakness Feature Proposal}
]
\small
\ttfamily

You are analyzing failures of an ObjectNav agent in procedurally
generated indoor environments.

Given successful and failed episodes, propose NEW numeric
environmental features that may predict policy failure and can
later be used to guide training environment generation.

\medskip
\textbf{Available episode fields}

targetCategory

scene:
\{houseArea, roomsTotal, numDoorsHouse, numTargetObjs,
numSameTypeRooms, numPlausibleDecoyRooms, targetRoomArea,
targetRoomDoors, targetRoomIsLeaf, targetInCorner,
targetDistToNearestDoor, occlusionDegree, losOnsetRange,
visibilityCoverage, startRoomType, startRoom, targetRoom,
roomPair, ...\}

planner:
\{shortestPath, roomHopsStartToTarget, doorsOnPath,
branchingBeforeTarget, decoyBranchesBeforeTarget,
doorwaySeesTarget, doorwaySeeingFrac,
pathBeforeFirstVisible\}

\medskip
\textbf{Successful episodes}

\{15 sampled successful episodes\}

\medskip
\textbf{Failed episodes}

\{15 sampled failed episodes\}

\medskip
\textbf{Output requirements}

1. Propose exactly 5 NEW candidate weakness features.

2. For each feature, write a Python function named
   feature\_*, taking one episode dictionary as input and
   returning a single numeric value.

3. Use only the episode fields listed above.

4. Prefer features corresponding to environmental properties
   that can be controlled during generation, such as geometry,
   layout, object placement, or their combinations.

5. Features may directly use one primitive or combine multiple
   primitives into ratios or interactions.

6. Guard against missing values.

Return all five functions in one Python code block.
\end{tcolorbox}

\subsection{Limitations}
We currently evaluate \method{} on ObjectNav in indoor 3D environments, using ProcTHOR for diagnosis and environment generation and HM3D for cross-domain evaluation. Extending the framework to additional navigation tasks, simulators, and real-world settings is a natural next step. Our current implementation also assumes access to measurable simulator primitives and controllable environment factors, which provide a clean testbed for studying weakness-guided generation. Finally, we focus on the top-ranked weakness in each adaptation round to keep the intervention targeted and interpretable. Future work could explore how to coordinate multiple complementary weaknesses within the same round.